\documentclass{article} 
\usepackage{iclr2021_conference,times}

\usepackage{amsmath,amsfonts,bm}

\def\eqref#1{equation~\ref{#1}}

\def\1{\bm{1}}

\DeclareMathAlphabet{\mathsfit}{\encodingdefault}{\sfdefault}{m}{sl}
\SetMathAlphabet{\mathsfit}{bold}{\encodingdefault}{\sfdefault}{bx}{n}

\usepackage{hyperref}
\usepackage{url}
\usepackage{graphicx}
\usepackage{multirow}
\usepackage{amsmath}
\usepackage{wrapfig}
\usepackage{booktabs} 
\usepackage{color}
\definecolor{light}{rgb}{0.3, 0.3, 0.3}
\def\light#1{{\color{light}#1}}

\usepackage{pifont}
\newcommand{\xmark}{\ding{55}}%

\title{GraphEBM: Molecular Graph Generation with Energy-Based Models}

\author{Meng Liu, Keqiang Yan, Bora Oztekin \& Shuiwang Ji \\
Department of Computer Science \& Engineering \\
Texas A\&M University \\
College Station, TX 77843, USA \\
\texttt{\{mengliu,keqiangyan,bora,sji\}@tamu.edu} \\
}

\iclrfinalcopy 
\begin{document}

\maketitle

\begin{abstract}
We note that most existing approaches for molecular graph generation fail to guarantee the intrinsic property of permutation invariance, resulting in unexpected bias in generative models. In this work, we propose GraphEBM to generate molecular graphs using energy-based models. In particular, we parameterize the energy function in a permutation invariant manner, thus making GraphEBM permutation invariant. We apply Langevin dynamics to train the energy function by approximately maximizing likelihood and generate samples with low energies. Furthermore, to generate molecules with a desirable property, we propose a simple yet effective strategy, which pushes down energies with flexible degrees according to the properties of corresponding molecules. Finally, we explore the use of GraphEBM for generating molecules with multiple objectives in a compositional manner. Comprehensive experimental results on random, goal-directed, and compositional generation tasks demonstrate the effectiveness of our proposed method.
\end{abstract}

\section{Introduction}

\label{sec:introduction}

A fundamental problem in drug discovery and material science is to find novel molecules with desirable properties. One way is to search in the chemical space based on molecular property prediction~\citep{gilmer2017neural,wu2018moleculenet,yang2019analyzing,stokes2020deep,wang2020advanced}. Recently, molecular graph generation has provided an alternative and promising way for this problem by directly generating desirable molecules, thus circumventing the expensive search of the chemical space. Despite intensive efforts recently, molecular graph generation remains challenging since the underlying graphs are discrete, irregular, and permutation invariant to node order. 


Existing approaches (Appendix~\ref{sec:molecule_generation}) have achieved promising success by generating molecular graphs based on various generative methods, including variational autoencoders (VAEs)~\citep{kingma2013auto}, generative adversarial networks (GANs)~\citep{goodfellow2014generative}, flow models~\citep{dinh2014nice,rezende2015variational} and recurrent neural networks (RNNs). However, as analyzed in Appendix~\ref{sec:molecule_generation}, most of them fail to preserve the intrinsic property of permutation invariance, which might yield different likelihoods for different permutations of the same graph.

Notably, energy-based models (EBMs)~\citep{lecun2006tutorial} can also be naturally used as generative models since data points near the underlying data manifold are assigned lower energies than other data points, which defines an unnormalized probability distribution over the data. Given a data point $x$, let $E_\theta(x) \in \mathbb{R}$ be the corresponding energy, where $\theta$ denotes the learnable parameters of the energy function. Then, the energy function defines a data distribution via the Boltzmann distribution as
\begin{equation}
\label{Eq:ebm_dist}
    p_\theta(x) = \frac{e^{- E_\theta(x)}}{Z(\theta)} \propto e^{- E_\theta(x)},
\end{equation}
where $Z(\theta)=\int e^{- E_\theta(x)}dx$ is the normalization constant and usually intractable. Related works on EBMs are reviewed in Appendix~\ref{app:ebm}.

In this work, we propose GraphEBM to generate molecular graphs with energy-based models. Since our parameterized energy function is permutation invariant, we can show that our GraphEBM preserves the permutation invariance property. We use Langevin dynamics~\citep{welling2011bayesian} to train the energy function by maximizing likelihood approximately and generate samples from the trained energy function. To our knowledge, our GraphEBM is the first energy-based model that can generate attributed molecular graphs. Furthermore, in order to generate molecules with a specific desirable property, we propose a novel, simple, and effective strategy to train our GraphEBM for goal-directed generation by pushing down energies with flexible degrees according to the property values of corresponding molecules. Significantly, we show that GraphEBM can generate molecules with multiple objectives in a compositional manner, which cannot be achieved by any existing methods. This provides a new and promising way for multi-objective molecule generation. Experimental results on random, goal-directed, and compositional generation tasks show that our proposed method is effective.


\section{The Proposed GraphEBM}
\label{sec:method}

In this section, we present GraphEBM by describing the internal structure of the parameterized energy function (Section~\ref{sec:energy_function}), showing that GraphEBM satisfies the desirable property of permutation invariance (Section~\ref{Sec:permutation_invirance}), and describing the training (Section~\ref{Sec:training}) and generation (Section~\ref{Sec:generation}) process of GraphEBM. Then, we introduce our proposed strategy for goal-directed generation based on GraphEBM (Section~\ref{Sec:goal_directed}). Finally, we explore the 
potential of compositional generation using GraphEBM (Section~\ref{Sec:compositional}).


Molecules can be naturally represented as graphs by considering atoms and bonds as nodes and edges, respectively. We formally represent a molecular graph as $\mathcal{G}=(X, A)$, where $X$ is the node feature matrix and $A$ is the adjacency tensor. Let $k$ be the number of nodes in the graph. $b$ and $c$ denote the number of possible types of nodes and edges, respectively. Then we have $X \in \{0,1\}^{k\times b}$ and $X_{(i,p)}=1$ if node $i$ belongs to type $p$. $A \in \{0,1\}^{k\times k\times c}$ and $A_{(i,j,q)}=1$ denotes that an edge with type $q$ exists between node $i$ and node $j$. Following~\citet{madhawa2019graphnvp} and~\citet{zang2020moflow}, we let $n$ denote the maximum number of atoms that a molecule has in a given dataset. We insert virtual nodes into molecular graphs that have less than $n$ nodes such that the dimensions of $X$ and $A$ keep the same for all molecules. Also, for any two nodes that are not connected in the molecule, we add a virtual edge between them. We can consider the virtual node and the virtual edge as an additional node type and edge type, respectively. Hence, for all molecules in a certain dataset, $X \in \{0,1\}^{n\times (b+1)}$ and $A \in \{0,1\}^{n\times n\times (c+1)}$.


\subsection{Parameterized Energy Function}
\label{sec:energy_function}
Following the above notations, the energy function for molecular graphs can be denoted as $E_\theta(X, A)$. Specifically, we model $E_\theta(X, A)$ by a graph neural network, where $\theta$ denotes parameters in the network. Many deep learning methods on graphs have been proposed and have achieved great success in many tasks, such as node classification~\citep{kipf2016semi,hamilton2017inductive,monti2017geometric,velivckovic2017graph,gao2018large,xu2018representation,klicpera2018predict,wu2019simplifying,liu2020towards,jin2020graph,chen2020simple,liu2020non,hu2020graphair}, graph classification~\citep{zhang2018end,ying2018hierarchical,maron2018invariant,xu2018powerful,gao2019graph,ma2019graph,Yuan2020StructPool:,you2020graph,gao2020topology}, and link prediction~\citep{zhang2018link,cai2020line}. In this work, we use a variant of relational graph convolutional networks (R-GCN)~\citep{schlichtkrull2018modeling} to learn the node representations since molecular graphs have categorical edge types. Formally, the layer-wise forward-propagation is defined as
\begin{equation}
\label{Eq:R-GCN}
    H^{\ell+1}=\sigma\left(\sum_{k=1}^{c+1}\left(A_{(:,:,k)}H^\ell W_k^\ell\right)\right).
\end{equation}
$A_{(:,:,k)}$ is the $k$-th channel of the adjacency tensor. $H^\ell \in \mathbb{R}^{n \times d_\ell}$ is the node representation matrix at layer $\ell$, where $d_\ell$ denotes the hidden dimension at layer $\ell$. $W_k^\ell \in \mathbb{R}^{d_\ell \times d_{\ell+1}}$ represents the trainable weight matrix for edge type $k$ at layer $\ell$. $\sigma(\cdot)$ denotes a non-linear activation function. The initial node representation matrix $H^0=X$. In each layer, message passing is conducted among the nodes independently for each type of edge. Then, the information is integrated together by a sum operator. We stack $L$ such layers. Hence, the final node representation matrix is $H^L \in \mathbb{R}^{n \times d}$, where $d$ is the hidden dimension. Then, the representation of the whole graph can be derived by a readout function. In this work, we use the sum operation to compute the graph-level representation $h_G$ as
\begin{equation}
\label{Eq:readout}
    h_G = \sum_{i=1}^n H_{(i,:)}^L \quad \in \mathbb{R}^d.
\end{equation}
Finally, the scalar energy associated with the molecular graph can be obtained by applying a transformation as
\begin{equation}
    E=h_G^TW \quad \in \mathbb{R},
\end{equation}
where $W \in \mathbb{R}^{d}$ is the trainable parameters.

\subsection{Permutation Invariance}
\label{Sec:permutation_invirance}

Permutation invariance is an intrinsic and desirable inductive bias for graph modeling. We note that our proposed GraphEBM satisfies this fundamental property due to our permutation invariant energy function. Specifically, each layer of our graph neural network in Eq.~(\ref{Eq:R-GCN}) is permutation equivariant. In addition, the readout operation in Eq.~(\ref{Eq:readout}) is permutation invariant. Therefore, our parameterized energy function is permutation invariant thus satisfying $E_\theta(X, A) = E_\theta(X^\pi, A^\pi)$, where $\pi$ denotes any permutation of node order. For simplicity, we use the superscript $\pi$ to denote that the corresponding matrix or tensor is arranged according to the node order given by $\pi$.
According to Eq.~(\ref{Eq:ebm_dist}), the energy function defines a distribution over data. Specifically, the likelihood is proportional to the negative exponential of the corresponding energy. Hence, we can further obtain $p_\theta(X, A)=p_\theta(X^\pi, A^\pi)$.
Thus, our GraphEBM can preserve permutation invariance by modeling graphs in a permutation invariant manner.

\subsection{Training}
\label{Sec:training}

Intuitively, a good energy function should assign lower energies to data points that correspond to real molecular graphs and higher energies to other data points. Hence, a straightforward idea is to train the parameterized energy function by maximizing the likelihood of real data defined in Eq.~(\ref{Eq:ebm_dist}). Let $p_D$ be the distribution of the real data. We can achieve maximum likelihood by minimizing the negative log likelihood of real data. Formally,
\begin{equation}
\label{Eq:loss_ml}
    \mathcal{L}_{ML}=\mathbb{E}_{(X,A)\sim p_D}\left[-\log p_\theta\left(X,A\right)\right],
\end{equation}
where $-\log p_\theta\left(X,A\right)=E_\theta(X, A)+\log Z(\theta)$ according to Eq.~(\ref{Eq:ebm_dist}). It has been shown~\citep{hinton2002training,Turner2005cd,song2021train} that the objective in Eq.~(\ref{Eq:loss_ml}) has the below gradient:
\begin{equation}
\label{Eq:gradient}
\nabla_\theta \mathcal{L}_{ML} = \mathbb{E}_{(X^\oplus,A^\oplus)\sim p_D} \left[\nabla_\theta E_\theta(X^\oplus, A^\oplus)\right]-\mathbb{E}_{(X^\odot,A^\odot)\sim p_\theta} \left[\nabla_\theta E_\theta(X^\odot, A^\odot)\right].
\end{equation}
As defined in Eq.~(\ref{Eq:ebm_dist}), $p_\theta$ is the distribution given by the energy function. Following~\citet{du2020improved}, we refer to $(X^\odot,A^\odot)$ as hallucinated samples. Obviously, this gradient pushes down the energies of positive samples $(X^\oplus,A^\oplus)$ and pushes up the energies of hallucinated samples $(X^\odot,A^\odot)$. However, sampling $(X^\odot,A^\odot)$ from $p_\theta$ is challenging, since $Z(\theta)$ in Eq.~(\ref{Eq:ebm_dist}) is intractable.

To overcome this issue, we follow~\citet{du2019implicit} to sample $(X^\odot,A^\odot)$ from an approximated $p_\theta$ using Langevin dynamics~\citep{welling2011bayesian}. Particularly, a sample $(X^\odot,A^\odot)$ is initialized randomly and refined iteratively by
\begin{equation}
\label{Eq:langevin_dynamics}
    X^k=X^{k-1}-\frac{\lambda}{2}\nabla_XE_\theta\left(X^{k-1},A^{k-1}\right)+w^k; \quad A^k=A^{k-1}-\frac{\lambda}{2}\nabla_AE_\theta\left(X^{k-1},A^{k-1}\right)+\eta^k,
\end{equation}
where $w^k$ and $\eta^k$ are added noise sampled from a Gaussian distribution $ \mathcal{N}(0,\sigma^2)$. $k$ denotes the iteration step, and $\frac{\lambda}{2}$ is the step size. As demonstrated by~\citet{welling2011bayesian}, the obtained samples $(X^k,A^k)$ approach samples from $p_\theta$ as $k\rightarrow\infty$ and $\frac{\lambda}{2}\rightarrow0$. In practice, we let $K$ denote the number of iteration steps of Langevin dynamics and use the resulting sample $(X^K,A^K)$ as $(X^\odot,A^\odot)$ in Eq.~(\ref{Eq:gradient}).

\begin{figure}[t]
    \begin{center}
        \includegraphics[width=\textwidth]{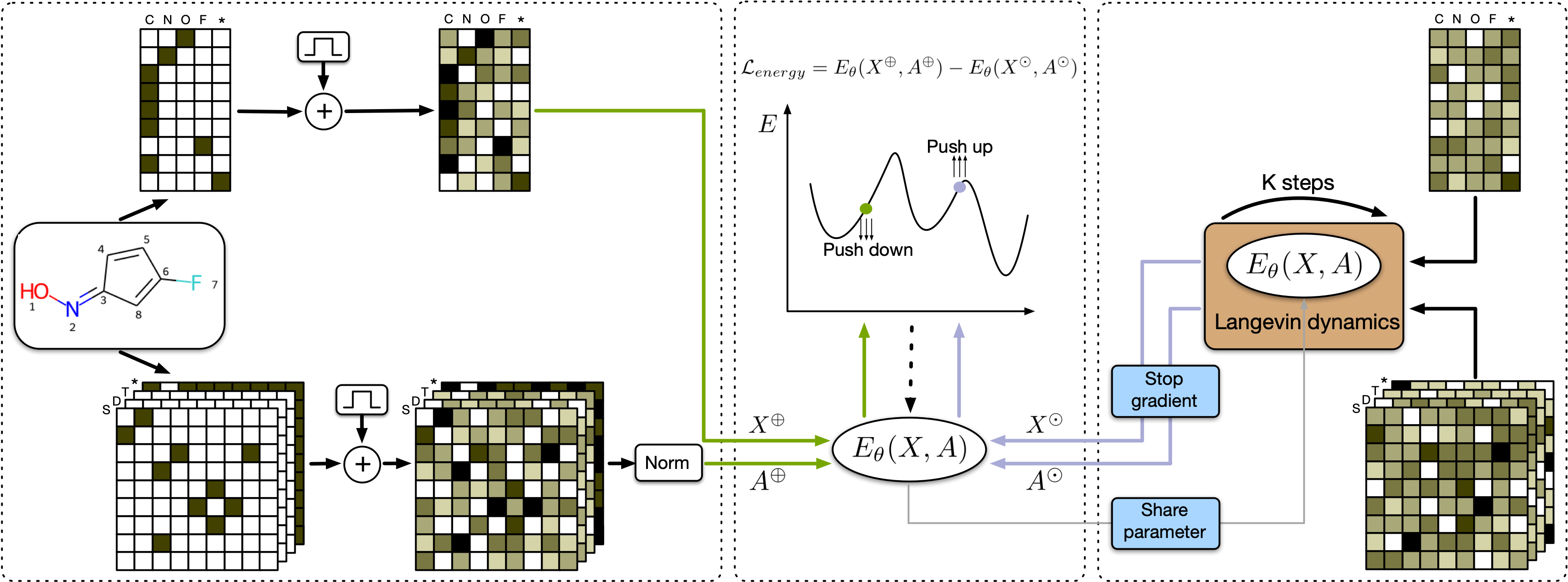}
    \end{center}
    \vspace{-0.1in}
    \caption{The training process of our GraphEBM. The left part and right part illustrate the processes of obtaining the positive sample and the hallucinated sample, respectively. The middle part shows the forward and backward propagation of the training process. Green and purple arrows represent the forward computation of energy value for the positive sample and the hallucinated sample, respectively. The black dashed arrow denotes the gradient backpropagation. In this example, $n=9$, $k=8$, $b=4$, and $c=3$. The annotations of node representation matrices denote the atom types, including carbon (C), nitrogen (N), oxygen (O), fluorine (F), and virtual atom ($\star$). Note that we remove hydrogen atoms, which is a common technique in the community. The annotations of adjacency tensor indicate the bond types, including single (S), double (D), triple (T), and virtual bond ($\star$).}
    \label{fig:graphebm_training}
\end{figure}

We illustrate the training process of our GraphEBM in Figure~\ref{fig:graphebm_training}. Since Langevin dynamics is for continuous data, we model the hallucinated samples by continuous format. For consistency, we can also use dequantization techniques~\citep{dinh2016density,kingma2018glow} to convert the discrete positive samples to continuous data by adding uniform noise, as shown in the left part of Figure~\ref{fig:graphebm_training}. The dequantization can be formally expressed as
\begin{equation}
        X'=X+tu, \quad u\sim[0,1)^{n\times(b+1)}; \quad A'=A+tu, \quad u\sim[0,1)^{n\times n \times(c+1)}.
\end{equation}
$t\in[0,1)$ is a scaling hyperparameter. We then apply a normalization to the adjacency tensor, which is a common step in modern graph neural networks~\citep{kipf2016semi}. Formally,
\begin{equation}
    A^\oplus_{(:,:,k)}=D^{-1}A'_{(:,:,k)}, \quad k=1,\cdots,c+1,
\end{equation}
where $D$ is the diagonal degree matrix in which $D_{(i,i)}=\sum_{j,k}A'_{(i,j,k)}$. We treat the above $A^\oplus$ and $X^\oplus=X'$ as the input for the energy function. In our case, each element of $X^\oplus$ is in $[0,1+t)$ and each element of $A^\oplus$ is in $[0,1)$.

Note that the above dequantization for positive samples is optional. This indicates that we can set $t=0$ and keep the positive data discrete since Langevin dynamics is only required for obtaining hallucinated samples. Applying dequantization to positive samples can be viewed as a data augmentation technique and we can easily convert the dequantized continuous data back to the original one-hot discrete data by simply applying the argmax operation.

To keep the same value range as $(X^\oplus,A^\oplus)$, the hallucinated sample $(X^\odot,A^\odot)$ is initialized as
\begin{equation}
\label{Eq:initialization}
    X^\odot\sim[0,1+t)^{n\times(b+1)}, \quad A^\odot\sim[0,1)^{n\times n \times(c+1)}.
\end{equation}
Then we apply $K$ steps of Langevin dynamics as Eq.~(\ref{Eq:langevin_dynamics}) to refine the sample, as illustrated in the right part of Figure~\ref{fig:graphebm_training}. After each step of refinement, we clamp the data to guarantee that the values are still in the desirable ranges.

As demonstrated in Eq.~(\ref{Eq:gradient}), the energies of positive samples are expected to be pushed down and the energies of hallucinated samples should be pushed up. Hence, to shape the energy function as expected, our loss function is defined as 
\begin{equation}
\label{Eq:loss_energy}
    \mathcal{L}_{energy}=E_\theta(X^\oplus, A^\oplus)-E_\theta(X^\odot, A^\odot).
\end{equation}
As shown in the middle part of Figure~\ref{fig:graphebm_training}, the gradient backpropagated from $\mathcal{L}_{energy}$ can update the parameters $\theta$, thus pushing the energy function $E_\theta(X,A)$ to approach our expected shape. Notably, the gradient from $\mathcal{L}_{energy}$ will not be propagated to the energy function used in Langevin dynamics. We apply parameter sharing to keep the energy function used in Langevin dynamics up-to-date.

To stabilize training, we also apply a regularization technique to the energy magnitudes. Specifically, we use the same regularization as~\citet{du2019implicit}. Formally,
\begin{equation}
    \mathcal{L}_{reg}=E_\theta(X^\oplus, A^\oplus)^2+E_\theta(X^\odot, A^\odot)^2.
\end{equation}
Hence, the total loss function is $\mathcal{L} = \mathcal{L}_{energy}+\alpha\mathcal{L}_{reg}$, where $\alpha \in \mathbb{R}$ is a hyperparameter.

\subsection{Generation}
\label{Sec:generation}

Let $E_{\theta^\star}(X,A)$ denote the trained energy function, where $\theta^\star$ represents the obtained parameters. Intuitively, if an energy function is well-shaped, the configurations with low energies should correspond to desirable molecular graphs. Hence, the generation process is to generate molecules based on the configurations $(X,A)$ that yield low energies.

An overview of the generation process is given in Figure~\ref{fig:graphebm_generation} in Appendix~\ref{app:gen}. The steps are as follows. First, we initialize a data point as in Eq.~(\ref{Eq:initialization}) and then apply $K$ steps of Langevin dynamics as in Eq.~(\ref{Eq:langevin_dynamics}) to obtain data points that have low energy. We denote the obtained configuration as $(X^\star,A^\star)$. Second, since molecular graphs are undirected, we make the adjacency tensor to be symmetric by using $A^\star+{A^\star}^T$ as the new adjacency tensor. Third, we convert the continuous data to discrete ones by applying the argmax operation in the dimensions of atom types and bond types. Finally, we use validity correction introduced by~\citet{zang2020moflow} to refine the corresponding molecule so that the valency constraint is satisfied.

\subsection{Goal-Directed Generation}
\label{Sec:goal_directed}

For drug discovery and material design, we also need to generate molecules with desirable chemical properties. This task is termed as goal-directed generation. As noted in Appendix~\ref{app:goal_directed}, it is not straightforward to apply existing strageties to our GraphEBM for goal-directed generation. To generate molecules with desirable chemical properties, we propose a novel, simple, and effective strategy to achieve goal-directed generation based on our GraphEBM. Our basic idea is to push down energies with flexible degrees according to the property values of corresponding molecules. If a molecule has a higher value of desirable property, we push down the corresponding energy harder. Formally, in goal-directed generation, the loss function defined in Eq.~(\ref{Eq:loss_energy}) becomes
\begin{equation}
\label{Eq:goal-directed}
    \mathcal{L}_{energy}=f(y)E_ \theta(X^\oplus, A^\oplus)-E_\theta(X^\odot, A^\odot),
\end{equation}
where $y\in[0,1]$ is the normalized property value and $f(y) \in \mathbb{R}$ determines the degree of the push down. We use $f(y)=1+e^y$ in this work. Thus, energies of molecules with higher property values are pushed down harder. After training, the generation process is the same as described in Section~\ref{Sec:generation}. Note that $f(y)$ could also be a learnable function and we leave this as future work.

\subsection{Compositional Generation}
\label{Sec:compositional}
In addition to single property constraints, it is commonly necessary to generate molecules with multiple property constraints in drug discovery~\citep{jin2020multi}. We observe that compositional generation with EBMs~\citep{hinton2002training}, which has been shown to be effective in the image domain~\citep{du2020compositional}, can be naturally applied to generate molecules with multiple constraints based on our GraphEBM. Thus, we investigate compositional generation in the graph domain.

Suppose we have two energy functions $E_{\theta_1^\star}(X,A)$ and $E_{\theta_2^\star}(X,A)$ trained towards two property goals respectively, as described in Section~\ref{Sec:goal_directed}. According to~\citet{hinton2002training} and~\citet{du2020compositional}, we can obtain a new energy function by summing the above two energy functions since the product of probabilities is equivalent to the sum of corresponding energies, according to Eq.~(\ref{Eq:ebm_dist}). Formally,
\begin{equation}
\label{Eq:compositional}
    E_{\theta^\star}(X,A)=E_{\theta_1^\star}(X,A)+E_{\theta_2^\star}(X,A).
\end{equation}
Then we can apply the generation process described in Section~\ref{Sec:generation} to $E_{\theta^\star}(X,A)$ to generate molecules towards multiple objectives in a compositional manner.

\section{Experiments}
\label{sec:experiments}

\begin{figure}[t]
\centering
\includegraphics[width=\textwidth]{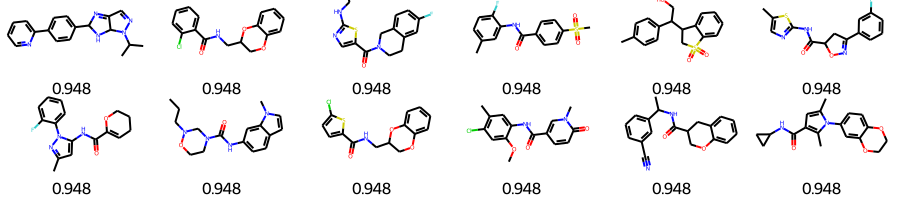}
\caption{Discovered examples with high QED scores.}
\label{fig:high_qed}
\end{figure}

We evaluate our proposed method for molecule generation under three settings: random generation, goal-directed generation, and compositional generation. We consider two widely used molecule datasets, QM9~\citep{ramakrishnan2014quantum} and ZINC250k~\citep{irwin2012zinc}. The details of the experimental setup for each setting are included in Appendix~\ref{app:exp}. Our implementation is included in DIG\footnote{https://github.com/divelab/DIG}~\citep{liu2021dig}, a library for graph deep learning research.

\begin{wrapfigure}{R}{7cm}
\vspace{-0.2in}
    \begin{center}
        \includegraphics[width=0.5\textwidth]{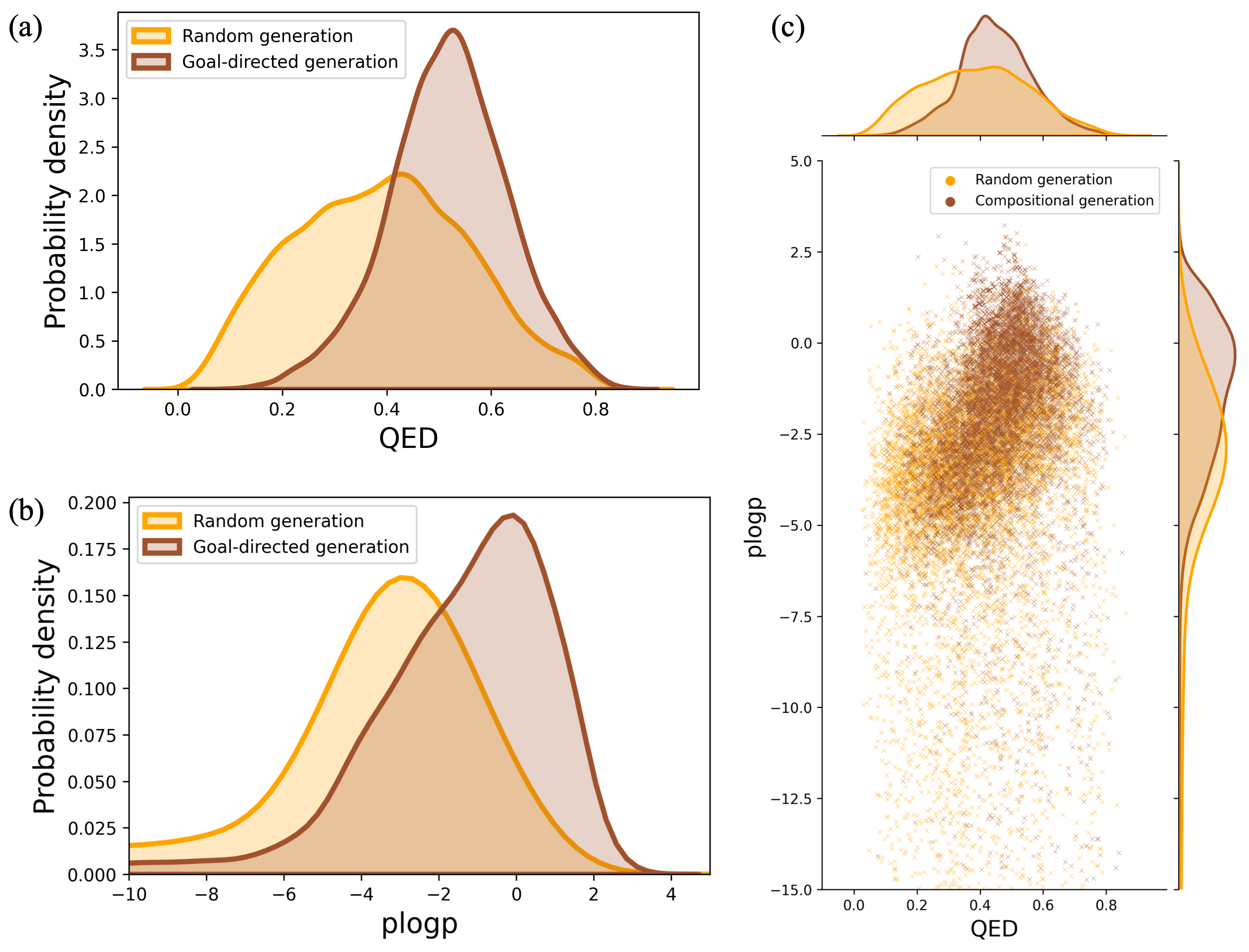}
    \end{center}
    \vspace{-0.2in}
    \caption{(a)\&(b) Comparison of QED and plogp distributions between goal-directed generation and random generation, respectively. (c) Comparison of distributions on QED and plogp between compositional generation and random generation.}
    \vspace{-0.3in}
    \label{fig:visualization_goal-directed}
\end{wrapfigure}

\textbf{Random generation}. The results on QM9 and ZINC250k are shown in Table~\ref{tab:gen_qm9} and Table~\ref{tab:gen_zinc250k} respectively in Appendix~\ref{app:exp_results}. We can observe that GraphEBM performs competitively with existing methods, which is significant considering that the study of EBMs is still in its early stage and GraphEBM is the first EBM for molecule generation. Generated samples are visualized in Figure~\ref{fig:mol} in Appendix~\ref{app:exp_results}, which further demonstrates that GraphEBM can generate non-trivial molecules.

To better understand the implicit generation through Langevin dynamics, we visualize this process for an example in Figure~\ref{fig:ld} in Appendix~\ref{app:exp_results}. We can observe that Langevin dynamics effectively refines the random initialized sample to approach a data point that corresponds to a realistic molecule.

\begin{wraptable}{R}{5.5cm}
\vspace{-0.7cm}
  \caption{Property optimization results.}
  \vskip 0.1in
  \label{tab:high_qed}
  \centering
  \resizebox{5.5cm}{!}{
    \begin{tabular}{lcccc}
    \toprule
    \textbf{Method} & \textbf{1st} & \textbf{2nd} & \textbf{3rd} & \textbf{4th}\\
    \midrule
    JT-VAE & $0.925$ & $0.911$ & $0.910$ & - \\
    GCPN & $0.948$ & $0.947$ & $0.946$ & - \\
    GraphAF & $0.948$ & $0.948$ & $0.947$ & $0.946$ \\
    MoFlow & $0.948$ & $0.948$ & $0.948$ & $0.948$ \\
    \midrule
    GraphEBM & $\textbf{0.948}$ & $\textbf{0.948}$ & $\textbf{0.948}$ & $\textbf{0.948}$ \\
    \bottomrule
  \end{tabular}
  }
  \vspace{-0.2in}
\end{wraptable}

\textbf{Goal-directed generation}. Figure~\ref{fig:visualization_goal-directed} (a) and (b) compare the property value distribution between goal-directed generation and random generation. It can be observed that goal-directed generation can generate more molecules with a high property value, indicating that our proposed strategy for goal-directed generation, which aims to assign lower energies to molecules with higher property values, is effective.

The property optimization results are shown in Table~\ref{tab:high_qed}. We observe that GraphEBM can find more novel molecules with the best QED score ($0.948$) than baselines. This strongly demonstrates the effectiveness of our proposed goal-directed generation method. Examples of discovered novel molecules with high QED scores are illustrated in Figure~\ref{fig:high_qed}

\begin{figure}[t]
\centering
\includegraphics[width=\textwidth]{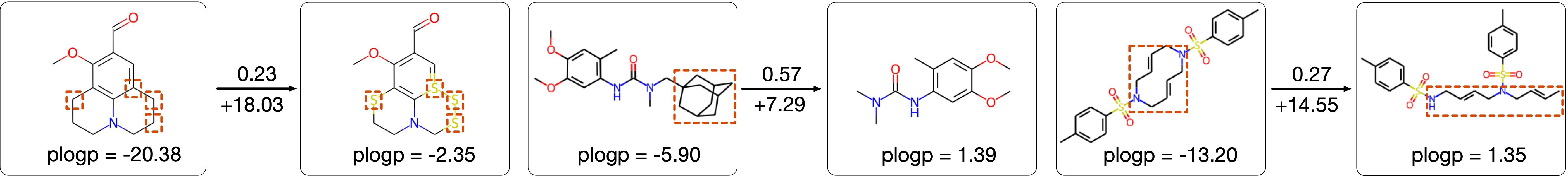}
\caption{Examples of constraint property optimization. The values above and below arrows denote the similarity scores and improvements, respectively. The modifications are highlighted with red rectangles.}
\label{fig:plogp_optimization}
\end{figure}

For constraint property optimization, as demonstrated in Table~\ref{tab:result_plogp}, GraphEBM can obtain higher property improvements over JT-VAE, GCPN, and MoFlow by significant margins. In terms of success rate, our GraphEBM is not as strong as the methods using reinforcement learning (\emph{i.e.}, GCPN, GraphAF), and achieves results comparable to JT-VAE and MoFlow. Although GraphAF performs better than GraphEBM, it can be observed from~\cite{shi2019graphaf} that GraphAF learns to improve plogp by simply adding long carbon chains, while our GraphEBM learns more advanced chemical knowledge. Several examples of constraint property optimization are shown in Figure~\ref{fig:plogp_optimization}. It is interesting that the modifications are interpretable to some degree. Specifically, in the first example, our model optimizes the plogp score with a remarkable margin of $18.03$ by replacing several carbon atoms with sulfur atoms, which could make the molecule more soluble in water, thus leading to a larger logP value. Additionally, plogp is highly related to the number of long cycles and synthetic accessibility. As shown in the second and third examples, our model improves the synthetic accessibility and reduces the number of long cycles by removing or breaking them. These facts indicate that our goal-directed generation method can explore the underlying chemical knowledge related to the corresponding property.

\begin{wraptable}{R}{8.5cm}
\vspace{-0.7cm}
    \caption{Constrained property optimization results. Imp., Sim., and Suc. denote Improvement, Similarity, and Success Rate, respectively.}
    \vskip 0.1in
    \begin{minipage}{0.6\columnwidth}
        \centering
        \label{tab:result_plogp}
        \resizebox{1\columnwidth}{!}{
            \begin{tabular}{lccc|ccc|ccc}
            \toprule
             & \multicolumn{3}{c}{JT-VAE} & \multicolumn{3}{c}{GCPN} & \multicolumn{3}{c}{GraphEBM} \\
            $\delta$ & \textbf{Imp.} & \textbf{Sim.} &\textbf{Suc.}  & \textbf{Imp.} & \textbf{Sim.} &\textbf{Suc.}  & \textbf{Imp.} & \textbf{Sim.} &\textbf{Suc.} \\
            \midrule
            $0.0$ &$1.91$ &$0.28$ &$98\%$ &$4.20$ &$0.32$ &$100\%$ &$\textbf{5.76}$ &$0.08$ &$98\%$   \\
            $0.2$ &$1.68$ &$0.33$ &$97\%$ &$\textbf{4.12}$ &$0.34$ &$100\%$ &$3.97$ &$0.35$ &$92\%$   \\
            $0.4$ &$0.84$ &$0.51$ &$84\%$ &$2.49$ &$0.47$ &$100\%$ &$\textbf{2.84}$ &$0.53$ &$88\%$   \\
            $0.6$ &$0.21$ &$0.69$ &$46\%$ &$0.79$ &$0.68$ &$100\%$ &$\textbf{1.52}$ &$0.68$ &$64\%$   \\
            \bottomrule
          \end{tabular}
        }
    \end{minipage} \\
    \begin{minipage}{0.6\columnwidth}
        \centering
        \resizebox{1\columnwidth}{!}{
            \begin{tabular}{lccc|ccc|ccc}
            \toprule
             & \multicolumn{3}{c}{GraphAF} & \multicolumn{3}{c}{MoFlow} & \multicolumn{3}{c}{GraphEBM} \\
            $\delta$ & \textbf{Imp.} & \textbf{Sim.} &\textbf{Suc.}  & \textbf{Imp.} & \textbf{Sim.} &\textbf{Suc.}  & \textbf{Imp.} & \textbf{Sim.} &\textbf{Suc.} \\
            \midrule
            $0.0$ &$13.13$ &$0.29$ &$100\%$ &$8.61$ &$0.30$ &$99\%$ &$\textbf{15.75}$ &$0.01$ &$99\%$   \\
            $0.2$ &$\textbf{11.90}$ &$0.33$ &$100\%$ &$7.06$ &$0.43$ &$97\%$ &$8.40$ &$0.35$ &$94\%$   \\
            $0.4$ &$\textbf{8.21}$ &$0.49$ &$100\%$ &$4.71$ &$0.61$ &$86\%$ &$4.95$ &$0.54$ &$79\%$   \\
            $0.6$ &$\textbf{4.98}$ &$0.66$ &$97\%$ &$2.10$ &$0.79$ &$58\%$ &$3.15$ &$0.67$ &$45\%$   \\
            \bottomrule
          \end{tabular}
        }
    \end{minipage}
    \vspace{-0.2in}
\end{wraptable}

\textbf{Compositional generation}. The comparison of the distributions on QED and plogp between compositional generation and random generation is illustrated in Figure~\ref{fig:visualization_goal-directed} (c). We can observe that compositional generation tends to generate more molecules with high QED and plogp scores. Additionally, the distribution of QED or plogp is similar to the corresponding distribution obtained by goal-directed generation towards a single objective (Figure~\ref{fig:visualization_goal-directed} (a) and (b)). These facts demonstrate that our GraphEBM is able to generate molecules towards multiple objectives in a composition manner, which provides a novel and promising way for multi-objective generation.

\section{Conclusion and Outlook}
\label{sec:conclusion}

In this paper, we propose GraphEBM, the first energy-based model for generating molecular graphs that preserves the intrinsic property of permutation invariance. We propose to flexibly push down energies for goal-directed generation and explore to generate molecules towards multiple objectives in a compositional manner, leading to a promising method for multi-objective generation. Experimental results demonstrate that our GraphEBM can generate realistic molecules and the proposals of goal-directed generation and compositional generation are effective and promising. Since EBMs have unique advantages and have rarely been explored for graph generation, we hope our exploratory work would open the door for future research in this area.

\section{Acknowledgments}

We thank Yilun Du and Chengxi Zang for their helpful discussions. This work was supported in part by National Science Foundation grant DBI-1922969.

\newpage
\bibliography{reference}
\bibliographystyle{iclr2021_conference}

\newpage
\appendix
\section*{Appendix}

\section{Related Work}
\label{sec:related_work}

\subsection{Molecular Graph Generation}
\label{sec:molecule_generation}

Since molecules can be represented as SMILES strings~\citep{weininger1988smiles}, early studies generate molecules based on SMILES strings, such as CVAE~\citep{gomez2018automatic}, GVAE~\citep{kusner2017grammar}, and SD-VAE~\citep{dai2018syntax}. Recent studies mostly represent and generate molecules as graphs~\citep{simonovsky2018graphvae,de2018molgan,madhawa2019graphnvp}. We can categorize existing molecular graph generation methods based on the underlying generative methods or the generation processes. Current molecular graph generation approaches can be grouped into four categories according to their underlying generative models, \emph{i.e.}, VAEs, GANs, flow models, and RNNs. They can also be classified into two primary types based on their generation processes; those are, sequential generation and one-shot generation. The sequential process generates nodes and edges in a sequential order by adding nodes and edges one by one. The one-shot process generates all nodes and edges at one time.

To facilitate comparison, we summarize existing methods in Table~\ref{tab:related_work}. We can observe that most of them fail to satisfy an intrinsic property of graphs; that is, permutation invariance. Specifically, a generative model should yield the same likelihood for different permutations of the same graph. Currently, permutation invariance remains to be a challenging goal to achieve. The sequential generation approaches have to choose a specific order of nodes, thus failing to preserve permutation invariance. Among the one-shot methods, ~\citet{bresson2019two} also use the specific node order given by the SMILES representation. GraphVAE and RVAE perform an approximate and expensive graph matching to train the VAE model, and they cannot achieve exact permutation invariance. MolGAN circumvents this issue by using a likelihood-free method. The recent one-shot flow methods have the potential to satisfy this property. However, GraphNVP, GRF, and MoFlow cannot preserve this property since the masking strategies in the coupling layers are sensitive to node order. An exception is GraphCNF, which achieves permutation invariance by assigning likelihood independent of node ordering via categorical normalizing flows.

In this work, we propose to develop energy-based models (EBMs)~\citep{lecun2006tutorial} for molecular graph generation. EBMs are a class of powerful methods for modeling richly structured data, but their use for graph generation has been under-explored.
We show that our method can achieve the desirable property of permutation invariance. Additionally, our method has the potential to generate molecules in a compositional manner towards multiple objectives, which cannot be achieved by any existing methods.

\begin{table}
	\caption{Summary and comparison of existing molecular graph generation methods.}
	\label{tab:related_work}
	\vskip 0.1in
	\centering
	\resizebox{\textwidth}{!}{
	\begin{tabular}{l|ccccc|cc|c|c}
		\toprule
		\multirow{2}{*}{\textbf{Method}} & \multicolumn{5}{c|}{\textbf{Generative method}} & \multicolumn{2}{c|}{\textbf{Generation process}} &\textbf{Permutation} &\textbf{Compositional}\\
		& \textit{VAE} &\textit{GAN} &\textit{Flow} &\textit{RNN} &\textit{EBM} &\textit{One-shot} &\textit{Sequential} & \textbf{invariance} &\textbf{generation} \\
		\midrule
	    GraphVAE~\citep{simonovsky2018graphvae} &\checkmark &- &- &- &- & \checkmark &- & \xmark &-\\
	    DeepGMG~\citep{li2018learning} &- &- &- &\checkmark &- &- &\checkmark & \xmark &-\\
	    CGVAE~\citep{liu2018constrained} &\checkmark &- &- &- &- & - &\checkmark & \xmark &-\\
	    MolGAN~\citep{de2018molgan} &- &\checkmark &- &- &- & \checkmark &- & - &-\\
	    RVAE~\citep{ma2018constrained} &\checkmark &- &- &- &- & \checkmark &- & \xmark &-\\
	    GCPN~\citep{you2018graph} &- &\checkmark &- &- &- & - &\checkmark & \xmark &-\\
	    JT-VAE~\citep{jin2018junction} &\checkmark &- &- &- &- & - &\checkmark & \xmark &-\\
	    MolecularRNN~\citep{popova2019molecularrnn} &- &- &- &\checkmark &- & - &\checkmark & \xmark &-\\
	    GraphNVP~\citep{madhawa2019graphnvp} &- &- &\checkmark &- &- & \checkmark &- & \xmark &-\\
	    \citet{bresson2019two} &\checkmark &- &- &- &- & \checkmark &- & \xmark &-\\
	    GRF~\citep{honda2019graph} &- &- &\checkmark &- &- & \checkmark &- & \xmark &-\\
	    GraphAF~\citep{shi2019graphaf} &- &- &\checkmark &- &- & - &\checkmark & \xmark &-\\
	    HierVAE~\citep{jin2020hierarchical} &\checkmark &- &- &- &- & - &\checkmark & \xmark &-\\
	    MoFlow~\citep{zang2020moflow} &- &- &\checkmark &- &- & \checkmark &- & \xmark &-\\
	    GraphCNF~\citep{lippe2020categorical} &- &- &\checkmark &- &- & \checkmark &- & \checkmark &-\\
	    \midrule
	    GraphEBM &- &- &- &- &\checkmark & \checkmark &- & \checkmark & \checkmark \\
		\bottomrule
	\end{tabular}
	}
\end{table}

\subsection{Energy-Based Models}
\label{app:ebm}
 Modeling variables by defining an unnormalized probability density has been explored for decades~\cite{hopfield1982neural,ackley1985learning,cipra1987introduction,dayan1995helmholtz,cipra1987introduction,zhu1998filters,hinton2012practical}. Such methods are known and unified as energy-based models (EBMs)~\citep{lecun2006tutorial} in machine learning. EBMs capture the dependencies of variables by assigning a scalar energy to each configuration of the variables with a learnable energy function. Given a trained EBM, inference is to find the configurations that yield low energies. Training an EBM aims at obtaining an energy function where observed configurations are associated with lower energies than unobserved ones. 

Currently, EBMs have been used as generative models in multiple domains, including images~\citep{xie2015learning,xie2016theory,du2019implicit,du2020compositional,du2020improved}, videos~\citep{xie2017synthesizing}, 3D objects~\citep{xie2018learning}, and point sets~\citep{xie2020generative}.


To date, EBMs have rarely been studied in the graph domain. ~\citet{liu2020gnnebm} attempt to generate graphs by building EBMs based on graph neural networks. \citet{niu2020permutation} model graphs using a score-based generative model~\citep{song2019generative}, a method that is similar to EBMs. However, these two methods can only generate graph structures, and it is not straightforward to use them on attributed graphs. In this work, we propose GraphEBM to generate attributed molecular graphs using EBMs. Hence, we consider GraphEBM to be the first energy-based model capable of generating attributed graphs.

\section{Generation Process of GraphEBM}
\label{app:gen}
\begin{figure}[t]
    \begin{center}
        \includegraphics[width=0.75\columnwidth]{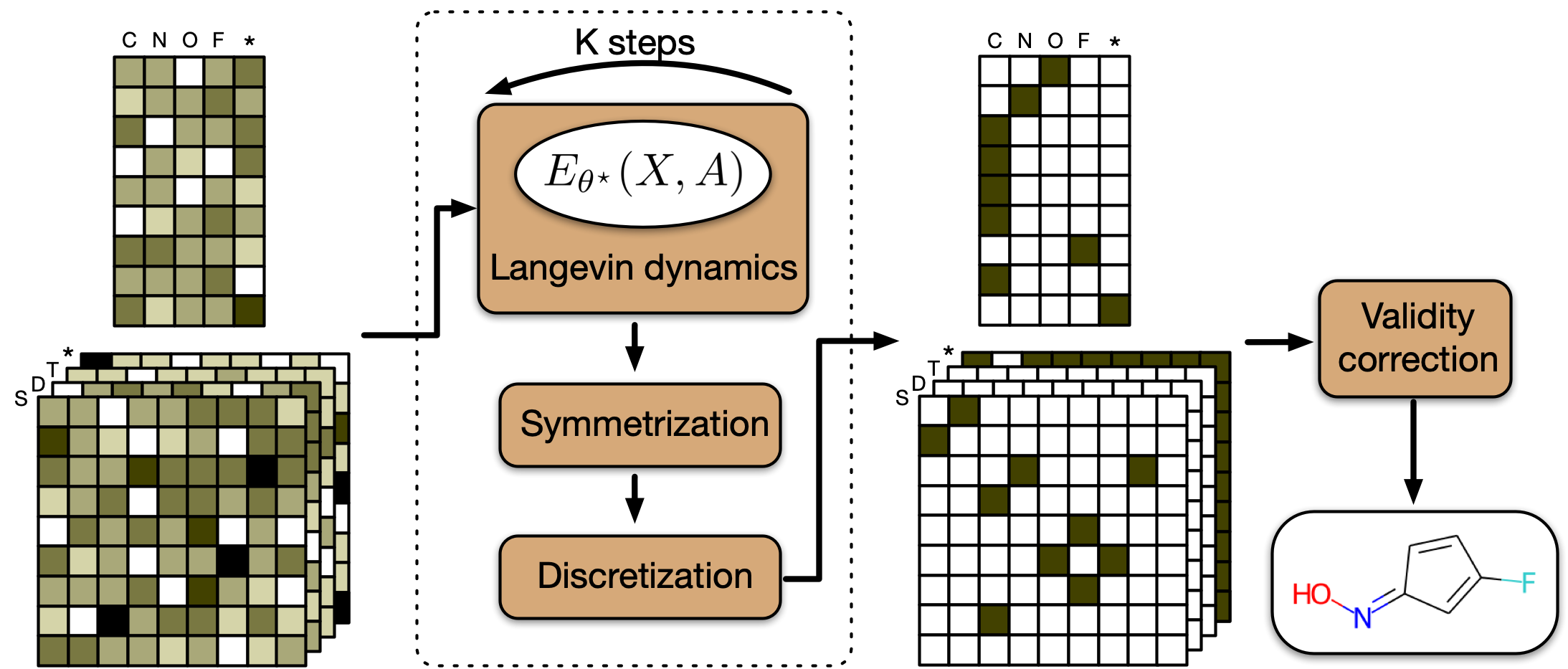}
    \end{center}
    \vspace{-0.2in}
    \caption{The generation process of our GraphEBM.}
    \label{fig:graphebm_generation}
\end{figure}

The generation process of our GraphEBM is illustrated in Figure~\ref{fig:graphebm_generation}.

\section{Existing Goal-Directed Generation Strategies}
\label{app:goal_directed}

There are mainly three approaches in the literature for goal-directed generation. First, this task can be modeled as a conditional generation problem, where the property value can be utilized as the condition~\citep{simonovsky2018graphvae}. Second, for methods using the latent space, a predictor can be applied to learn the property value from the latent representation~\citep{gomez2018automatic}. Third, reinforcement learning can be used to optimize the properties of generated molecules~\citep{you2018graph}. However, it is not straightforward to apply these methods to our GraphEBM for goal-directed generation since GraphEBM generates molecules implicitly using Langevin dynamics and no latent space exists. Also, using EBMs for generation that is conditional on continuous conditions is rarely studied by the community. Hence, it remains challenging to apply EBMs for goal-directed generation.

\section{Experimental Setup}
\label{app:exp}

\textbf{Dataset}. QM9 consists of $134$k organic molecules and the maximum number of atoms is $9$. It contains $4$ atom types and $3$ bond types. ZINC250k has $250$k drug-like molecules and the maximum number of atoms is $38$. It includes $9$ atom types and $3$ edge types.

\textbf{Implementation details}. We kekulize molecules and remove their hydrogen atoms using RDKit~\citep{landrum2006rdkit}. In our parameterized energy function, we adopt a network of $L=3$ layers with hidden dimension $d=64$. We use Swish as the activation function. We set $\alpha=1$ in the loss function and the standard variance $\sigma=0.005$ in the gaussian noise. For training, we tune the following hyperparameters: the scale $t$ of uniform noise $\in$ [0, 1], the sample step $K$ of Langevin dynamics $\in$ [30, 300], and the step size $\frac{\lambda}{2}$ $\in$ [10, 50]. All models are trained for up to $20$ epochs with a learning rate of $0.0001$ and a batch size of $128$. It is well known that it is difficult to train EBMs. We follow the techniques adopted in~\citet{du2019implicit} to stabilize the training process. Specifically, we add spectral normalization~\citep{miyato2018spectral} to all layers of the network. In addition, we clip the gradient used in Langevin dynamics so that its value magnitude can be less than $0.01$. GraphEBM is implemented with PyTorch~\citep{paszke2017automatic}.

\textbf{Random generation}. We evaluate the ability of our proposed GraphEBM to model and generate molecules. We consider most methods reviewed in Section~\ref{sec:molecule_generation} as baselines. The following widely used metrics are adopted. \textit{Validity} is the percentage of chemically valid molecules among all generated molecules. \textit{Uniqueness} denotes the percentage of unique molecules among all valid molecules. \textit{Novelty} corresponds to the percentage of generated valid molecules that are not present in the training set. The metrics are computed on $10,000$ randomly generated molecules. Results averaged over $5$ runs are reported.


\textbf{Goal-directed generation}. To empirically show the effectiveness of our goal-directed generation method proposed in Section~\ref{Sec:goal_directed}, we train models on ZINC250k accordingly and compare the distribution of the property score between goal-directed generated molecules and random generated molecules. We consider two chemical properties, including Quantitative Estimate of Druglikeness (QED)~\citep{bickerton2012quantifying} and penalized logP (plogp), which is the water-octanol partition coefficient penalized by the number of long cycles and synthetic accessibility.

We further verify the effectiveness of our proposed goal-directed generation method by performing molecule optimization, including property optimization and constrained property optimization. Property optimization aims at generating novel molecules with high QED scores. We directly use the model trained for goal-directed generation and leverage the molecules in the training set as initialization for Langevin dynamics, following prior works~\citep{jin2018junction,zang2020moflow}. We report the highest QED scores and the corresponding novel molecules discovered by our method. For constrained property optimization, given a molecule $m$, our task is to obtain a new molecule $m'$ that has a better desired chemical property with the molecular similarity $sim(m, m') \geq \delta$ for some threshold $\delta$. We adopt Tanimoto similarity of Morgan fingerprint~\citep{rogers2010extended} to measure the similarity between molecules. We find that there are two different settings in baselines. JT-VAE and GCPN choose $800$ molecules with the lowest plogp scores in the test set and use them as initialization, while GraphAF and MoFlow choose from the training set. We report our results on both of these two settings for extensive comparisons.

\textbf{Compositional generation}. As investigated in Section~\ref{Sec:compositional}, our GraphEBM has the potential to conduct compositional generation towards multiple objectives. To verify this, we combine the two energy functions obtained in goal-directed generation experiments, as formulated in Eq.~(\ref{Eq:compositional}). Then we apply the generation process described in Section~\ref{Sec:generation} to the resulting energy function to generate molecules.

\section{Experimental Results}
\label{app:exp_results}

\begin{figure}[t]
    \begin{center}
        \includegraphics[width=0.75\columnwidth]{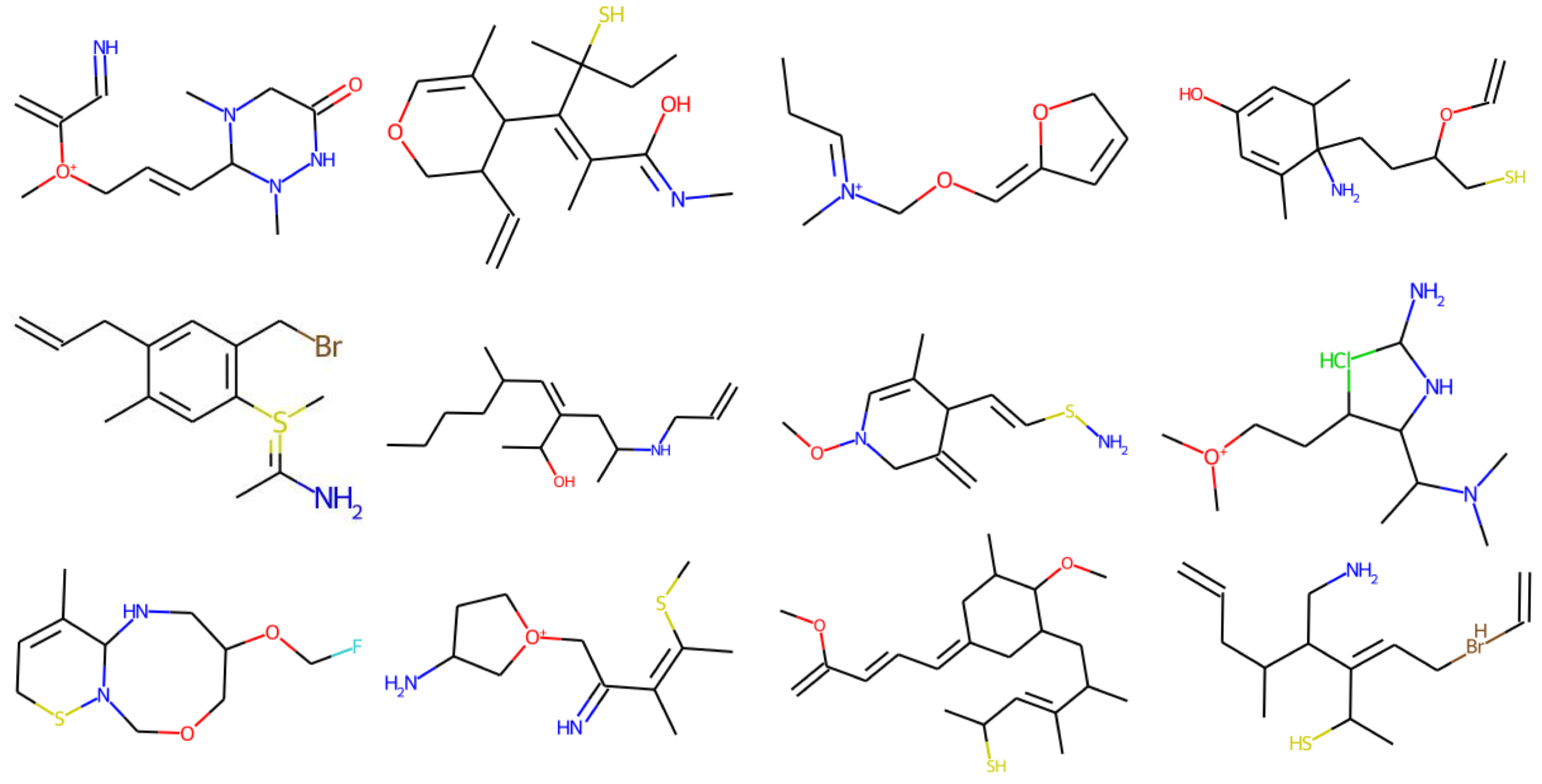}
    \end{center}\vspace{-0.6cm}
    \caption{Visualization of molecules generated by GraphEBM.}
    \label{fig:mol}
\end{figure}

\begin{figure}[h!]
    \begin{center}
        \includegraphics[width=0.75\columnwidth]{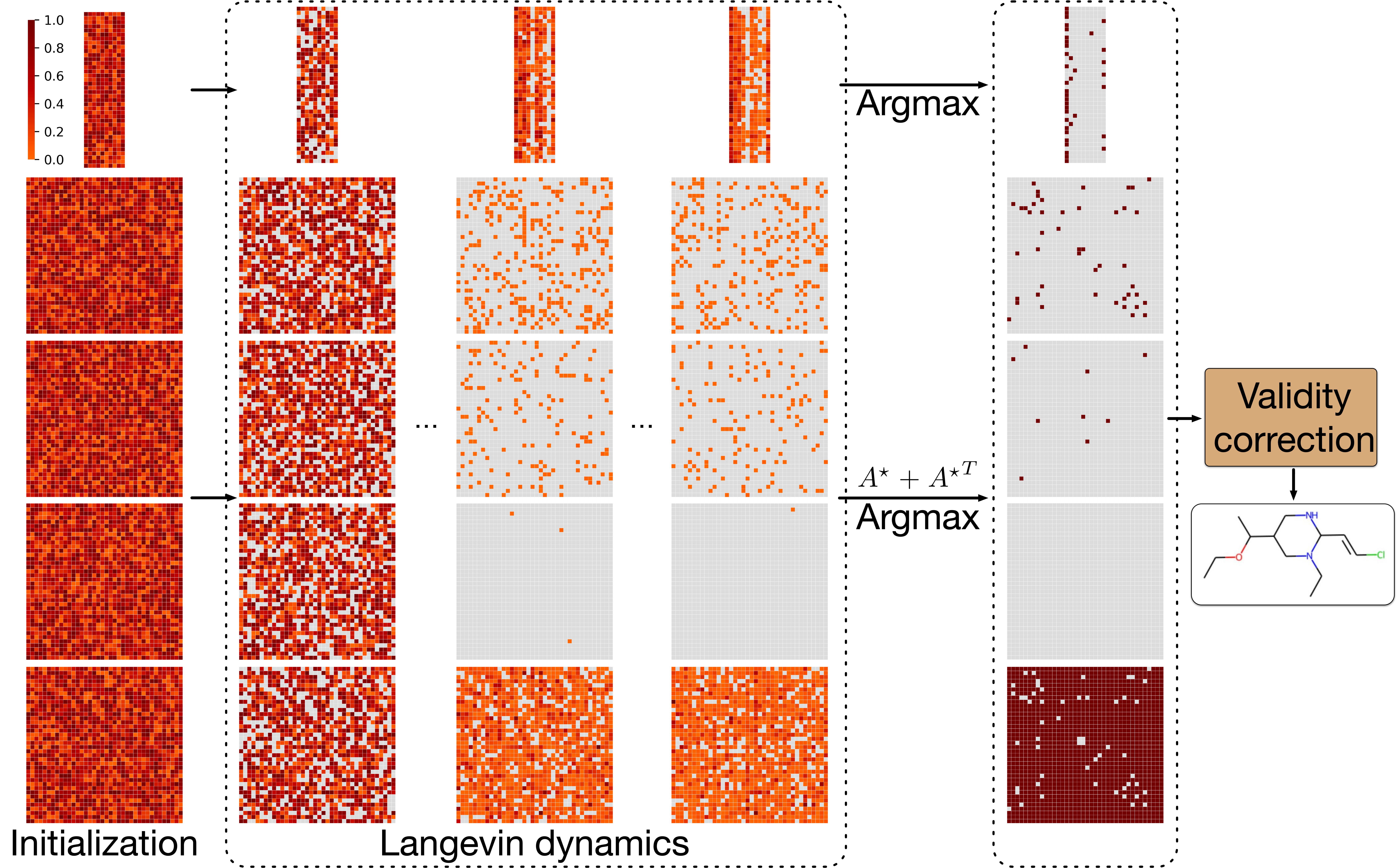}
    \end{center}
    \vspace{-0.2in}
    \caption{Visualization of the implicit generation process of our GraphEBM. The first row denotes atom matrices and the remaining rows represent fours channels of adjacency tensors, corresponding to single, double, triple, and virtual bonds. For better visual results, each atom matrix and adjacency tensor is normalized by dividing by its maximum value.}
    \label{fig:ld}
\end{figure}

The generation performance on QM9 and ZINC250k is given in Table~\ref{tab:gen_qm9} and Table~\ref{tab:gen_zinc250k}. The generated molecule samples are visualized in Figure~\ref{fig:mol}. The visualization of the implicit generation process of our GraphEBM is shown in Figure~\ref{fig:ld}.

\begin{table}[h]
\begin{minipage}{0.48\columnwidth}
	\caption{Generation performance on QM9. The results of CVAE and GVAE are obtained from~\citet{simonovsky2018graphvae}. The result of MoFlow is obtained by evaluating its public trained model. All other results are from their original papers.}
	\label{tab:gen_qm9}
	\vskip 0.1in
	\centering
	\resizebox{1\columnwidth}{!}{
	\begin{tabular}{lccc}
		\toprule
		\textbf{Method} & \textbf{Validity}(\%) & \textbf{Uniqueness}(\%) &\textbf{Novelty}(\%) \\
		\midrule
	    CVAE &$10.30$ &$67.50$ &$90.00$ \\
	    GVAE &$60.20$ &$9.30$ &$80.90$ \\
	    GraphVAE &$55.70$ &$76.00$ &$61.60$ \\
	    RVAE &$96.60$ &- &$97.50$ \\
	    MolGAN &$98.10$ &$10.40$ &$94.20$ \\
	    GraphNVP &$83.10\scriptstyle{{\light{\pm0.50}}}$ & $99.20\scriptstyle{{\light{\pm0.30}}}$ & $58.20\scriptstyle{{\light{\pm1.90}}}$ \\
	    GRF &$84.50\scriptstyle{{\light{\pm0.70}}}$ & $66.00\scriptstyle{{\light{\pm1.14}}}$ & $58.60\scriptstyle{{\light{\pm0.82}}}$ \\
	    GraphAF &$100.00$ &$94.51$ &$88.83$ \\
	    MoFlow &$100.00\scriptstyle{{\light{\pm0.00}}}$ & $98.53\scriptstyle{{\light{\pm0.14}}}$ & $96.04\scriptstyle{{\light{\pm0.10}}}$ \\
	    \midrule
	    \textbf{GraphEBM} &$100.00\scriptstyle{{\light{\pm0.00}}}$ & $97.90\scriptstyle{{\light{\pm0.14}}}$ & $97.01\scriptstyle{{\light{\pm0.17}}}$ \\
		\bottomrule
	\end{tabular}
	}
\end{minipage}
\hspace{0.5cm}
\begin{minipage}{0.48\columnwidth}
	\caption{Generation performance on ZINC250k. The results of GCPN and JT-VAE are obtained from~\citet{shi2019graphaf}. The result of MoFlow is obtained by evaluating its public trained model. All other results are from their original papers.}
	\label{tab:gen_zinc250k}
	\vskip 0.1in
	\centering
	\resizebox{1\columnwidth}{!}{
	\begin{tabular}{lccc}
		\toprule
		\textbf{Method} & \textbf{Validity}(\%) & \textbf{Uniqueness}(\%) &\textbf{Novelty}(\%) \\
		\midrule
	    GCPN &$100.00$ &$99.97$ &$100.00$ \\
	    JT-VAE &$100.00$ &$100.00$ &$100.00$ \\
	    MolecularRNN &$100.00$ &$99.89$ &$100.00$ \\
        GraphNVP &$42.60\scriptstyle{{\light{\pm1.60}}}$ & $94.80\scriptstyle{{\light{\pm0.60}}}$ & $100.00\scriptstyle{{\light{\pm0.00}}}$ \\
        GRF &$73.40\scriptstyle{{\light{\pm0.62}}}$ & $53.7\scriptstyle{{\light{\pm2.13}}}$ & $100.00\scriptstyle{{\light{\pm0.00}}}$ \\
	    GraphAF &$100.00$ &$99.10$ &$100.00$ \\
	    MoFlow &$100.00\scriptstyle{{\light{\pm0.00}}}$ & $99.99\scriptstyle{{\light{\pm0.01}}}$ & $100.00\scriptstyle{{\light{\pm0.00}}}$ \\
	    \midrule
	    \textbf{GraphEBM} &$99.96\scriptstyle{{\light{\pm0.02}}}$ & $98.79\scriptstyle{{\light{\pm0.15}}}$ & $100.00\scriptstyle{{\light{\pm0.00}}}$ \\
		\bottomrule
	\end{tabular}
	}
\end{minipage}
\end{table}

\end{document}